\documentclass[a4paper,twoside]{article}

\usepackage{epsfig}
\usepackage{subcaption}
\usepackage{calc}
\usepackage{amssymb}
\usepackage{amstext}
\usepackage{amsmath}
\usepackage{amsthm}
\usepackage{multicol}
\usepackage{pslatex}
\usepackage{apalike}
\usepackage{SCITEPRESS}     

\begin{document}

\title{Stabilizing GANs with Soft Octave Convolutions}

\author{\authorname{Ricard Durall\sup{1,2,3}, Franz-Josef Pfreundt\sup{1} and Janis Keuper\sup{1,4}}
\affiliation{\sup{1}Fraunhofer ITWM, Germany}
\affiliation{\sup{2}IWR, University of Heidelberg, Germany}
\affiliation{\sup{3}Fraunhofer Center Machine Learning, Germany}
\affiliation{\sup{4}Institute for Machine Learning and Analytics, Offenburg University, Germany}
}


\abstract{ Motivated by recently published methods using frequency decompositions of convolutions (e.g. Octave Convolutions), we propose a novel convolution scheme to stabilize the training and reduce the likelihood of a mode collapse. The basic idea of our approach is to split convolutional filters into additive high and low frequency parts, while shifting weight updates from low to high during the training. Intuitively, this method forces GANs to learn low frequency coarse image structures before descending into fine (high frequency) details. We also show, that the use of the proposed soft octave convolutions reduces common artifacts in the frequency domain of generated images. Our approach is orthogonal and complementary to existing stabilization methods and can simply be plugged into any CNN based GAN architecture. Experiments on the CelebA dataset show the effectiveness of the proposed method. }

\onecolumn \maketitle \normalsize \setcounter{footnote}{0} \vfill

\section{\uppercase{Introduction}}
\label{sec:introduction}

\let\thefootnote\relax\footnotetext{Accepted in VISAPP 2021}

\noindent In recent years, unsupervised learning has received a lot of attention in computer vision applications. In particular, learning generative models from large and diverse datasets has been a very active area of research. Generative Adversarial Networks (GANs)\,\cite{goodfellow2014generative} has risen as one of the main techniques that produces state-of-the-art results at generating realistic and sharp images. Unlike other generative methods\,\cite{kingma2013auto,oord2016pixel} that explicitly model maximum likelihood, GAN provides an attractive alternative that allows to model the density implicitly. Basically, it consists of training a generator and discriminator model in an adversarial game, such that the generator learns to produce samples from the data distribution. Nonetheless, despite their success, GANs often have an unstable training behaviour of which there is little to no theory explaining it. This makes it extremely hard to predict plausible results in new GAN experiments or to employ them in new domains. Consequently, their applicability is often drastically limited to a controlled and well-defined environment. In the literature, we encounter many current papers dedicated to finding heuristically stable architectures\,\cite{radford2015unsupervised,karras2017progressive,brock2018large,lin2019coco}, loss functions\,\cite{mao2017least,arjovsky2017wasserstein,gulrajani2017improved} or regularization strategies\,\cite{miyato2018spectral,durall2020watch}.

All the aforementioned generative methods have a common core building block: convolutions. In other words, they all are based on convolutional neural networks (CNN). That means that their architecture consist mostly of sets of stacked convolutional layers. Recent efforts have focused on improving these layers by reducing their inherent redundancy in density of parameters and in the amount of channel dimension of feature maps\,\cite{han2016dsd,luo2017thinet,chollet2017xception,xie2017aggregated,ke2017multigrid,chen2019drop}. These works analyse standard convolutional layers and their behaviour in detail. Basically, these layers are designed to detect local conjunctions of features from the previous layer and mapping their appearance to a feature map, which have always the same spatial resolution. However, natural images can be factorized into a low frequency signal that captures the global layout and coarse structure, and a high frequency part that captures fine details. Attracted by the idea of having feature maps with different resolutions and breaking with standard convolutional layers, some works\,\cite{ke2017multigrid,chen2019drop} have built schemes, on top of standard CNNs architecture, that have access to different frequency content within the same feature map.

In this paper, we propose to replace standard convolutions from the architecture of GANs with novel soft octave convolutions. This replacement will have almost no impact on the architecture since octave convolutions are orthogonal and complementary to existing methods that also focus on improving CNN topology. We apply our model to the CelebA \cite{liu2015deep} dataset and demonstrate that, by simply substituting the convolutional layers, we can consistently improve the performances leading to a more stable training with less probability of mode collapses. Overall, our contributions are summarized as follows:

\begin{itemize}
\item  We introduce a novel and generalizable convolution scheme for generative adversarial networks: the soft octave convolution. 

\item  Our analysis shows that, using soft octave convolutions leads to more stable training runs and less frequency domain artifacts in generated images. 

\item We evaluate our approach by embedding the soft octave convolutions into different GAN architectures and provide both quantitative and qualitative results on the CelebA dataset.
\end{itemize}

\section{\uppercase{Related Work}}

\noindent Most of the deep learning approaches in computer vision are based on standard CNNs. They have heavily contributed in semantic image understanding tasks including the aforementioned works and references therein. In this work, we look at image generation techniques and we briefly review the seminal work in that direction. In particular, we focus our attention on a set of well-known GANs and the impact of alternative convolutional layers on these models.

\subsection{Generative Adversarial Networks}

\noindent The goal of generative models is to match real data distribution $p_{\mathrm{data}}$ with generated data distribution $p_{\mathrm{g}}$. Thus, minimizing differences between two distributions is a crucial point for training generative models. Goodfellow\textit{ et al.} introduced an adversarial framework (GAN)\,\cite{goodfellow2014generative} which is capable of learning deep generative models by minimizing the Jensen-Shannon Divergence between $p_{\mathrm{data}}$ and $p_{\mathrm{g}}$. This optimization problem can be described as a minmax game between the generator $G$, which learns how to generate samples which resemble real data, and a discriminator $D$, which learns to discriminate between real and fake data. Throughout this process, $G$ indirectly learns how to model $p_{\mathrm{data}}$ by taking samples $z$ from a fixed distribution $p_{\mathrm{z}}$ (e.g. Gaussian) and forcing the generated samples  $G(z)$ to match $p_{\mathrm{g}}$. The objective loss function is defined as

\begin{align}
\begin{split}
	\min_{G} \max_{D} \mathcal{L}(D,G) =\,& \mathbb{E}_{\mathrm{\mathbf{x}} \sim p_{\mathrm{data}}} \left[ \log \left(D(\mathrm{\mathbf{x}})\right) \right] \,+ \\
	& \mathbb{E}_{\mathrm{\mathbf{z}} \sim p_{\mathrm{z}}}[\log(1-D(G(\mathrm{\mathbf{z}}))].
\end{split}
\end{align}
\vspace{1px}

\noindent \textbf{Deep Convolutional GAN.} Deep Convolutional GAN (DCGAN)\,\cite{radford2015unsupervised} is one of the popular and successful network topology designs for GAN that in a certain way achieves a consistently stability during training. It is a direct extension of the GAN described above, except that it is mainly composed of convolution and transposed-convolution layers without max pooling or fully connected layers in both discriminator and generator.\\

\noindent \textbf{Least-Squares GAN.} Least-Squares GAN (LSGAN)\,\cite{mao2017least} also tries to minimize Pearson $X^{2}$ divergence between the real and the generated distribution. The standard GAN uses a sigmoid cross entropy loss for the discriminator to determine whether its input comes from $p_{\mathrm{data}}$ and $p_{\mathrm{g}}$. Nonetheless, this loss has an important drawback. Given a generated sample is classified as real by the discriminator, then there would be no apparent reason for the generator to be updated even though the generated sample is located far from the real data distribution. In other words, sigmoid cross entropy loss can barely push such generated samples towards real data distribution since its classification role has been achieved. Motivated by this phenomenon, LSGAN replaces a sigmoid cross entropy loss with a least square loss, which directly penalizes fake samples by moving them close to the real data distribution.\\

\noindent \textbf{Wasserstein GAN.} Wasserstein GAN (WGAN)\,\cite{arjovsky2017wasserstein} suggests the Earth-Mover (EM) distance, which is also called the Wasserstein distance, as a measure of the discrepancy between the two distributions. The benefit of the EM distance over other metrics is that it is a more sensible objective function when learning distributions with the support of a low-dimensional manifold. EM distance is continuous and differentiable almost everywhere under Lipschitz condition, which standard feed-forward neural networks satisfy. In order to enforce such a condition, weight clipping is used on each neural network layer. Its main idea is to clamp the weight to a small range, so that the Lipschitz continuity is guaranteed. Finally, since EM distance is intractable, it is converted in to a tractable equation via Kantorovich-Rubinstein duality with the Lipschitz function.

\subsection{Convolutional Layers}

\noindent Standard convolutional layers are designed to detect local conjunctions of features from the previous layer and to project their appearance to a feature map which does not vary its spatial resolution at any time. 
Nevertheless, in accordance with the spatial-frequency model\,\cite{campbell1968application,devalois1990spatial}, natural images can be factorized into a low frequency signal that captures the global layout and coarse structure, and a high frequency signal that captures fine details.
Attracted by the idea of having feature maps with different resolution, recent works tried to leverage frequency decompositions for  deep learning approaches. For example, \cite{ke2017multigrid,chen2019drop} have built architectures on top of standard CNNs, that have access to different frequency content. \cite{ke2017multigrid} suggested a multigrid architecture, that has the intention of wiring cross-scale connections into network structure at the lowest level. In order to create such a topology, every convolutional filter extends spatially within grids ($h$,$w$), across grids multiple scales ($s$) within a pyramid, and over corresponding feature channels ($c$). Building in this fashion, a combination of pyramids across the architecture ($h$,$w$,$s$,$c$). 

\subsection{Octave Convolutions}
\begin{figure}[!t]
\begin{center}
   \includegraphics[width=\linewidth]{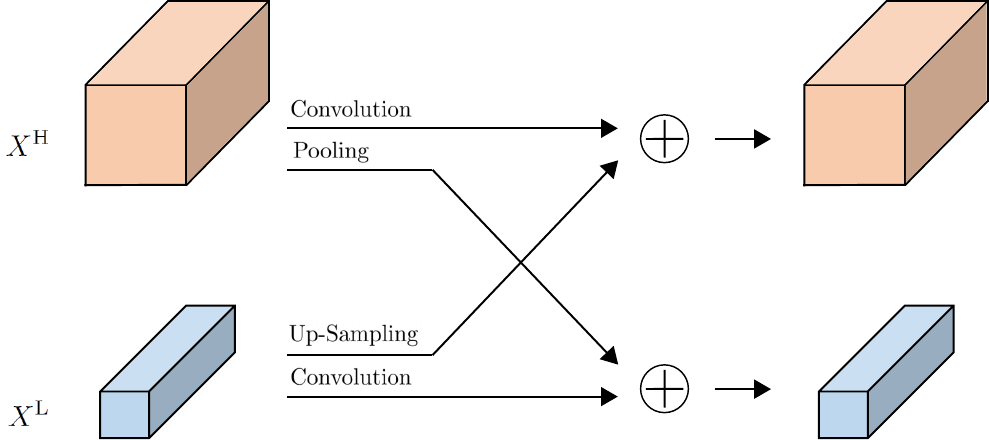}
\end{center}
   \caption{Original formulation of the octave convolution as introduced by \cite{chen2019drop}. Inputs $X$ to convolutional layers are separated into  $X = \{X^{\mathrm{H}}, X^{\mathrm{L}}\}$ along the channel domain. Each layer then computes convolutions on high ($X^{\mathrm{H}}$) and low ($X^{\mathrm{L}}$) frequency parts which are then recombined in the output.}
\label{fig:octave}
\end{figure}

\noindent The original approach towards octave convolutions has been introduced by \cite{chen2019drop}.
Given the input feature tensor of a convolutional layer $X \in \mathbb{R}^{c \times h \times w}$ with channels $c$ and spacial resolutions in height $h$ and width $w$, \cite{chen2019drop} suggested to factorize it along channel dimension into two groups, one for low frequencies and one for high frequencies $X = \{X^{\mathrm{H}}, X^{\mathrm{L}}\}$ (see Fig.\,\ref{fig:octave} for details).  The authors argued, that the subset of the feature maps that capture spatially low frequency changes contains spatially redundant information. In order to reduce the spatial redundancy, they introduced the octave feature representation, which corresponds to a division of the spatial dimensions by 2 for some of the feature maps.

\begin{figure}[!t]
\begin{center}
   \includegraphics[width=\linewidth]{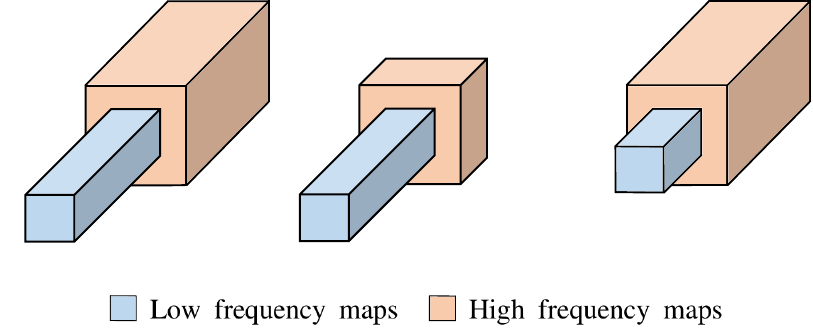}
\end{center}
   \caption{In the original octave convolution, the hyper-parameter $\alpha$ sets the rate of how a fixed number of convolution filters is split into high and low frequency maps. Left to right: $\alpha=0.5, 0.25, 0.75$. Note: since $\alpha$ changes the network topology, it can not  be changed during training.}
\label{fig:octave_2}
\end{figure}

In order to control the factorization into high- and low frequency parts, \cite{chen2019drop} introduced the hyper-parameter $\alpha \in [0,1]$ (see Fig.\,\ref{fig:octave_2})

\begin{align}
\begin{split}
	X^{\mathrm{L}} \in \mathbb{R}^{\alpha c \times \frac{h}{2} \times \frac{w}{2}} \quad \mathrm{and} \quad X^{\mathrm{H}} \in \mathbb{R}^{(1-\alpha)c \times h \times w}.
\end{split}
\end{align}
\vspace{1px}

However, this formulation has a major drawback in practice: $\alpha$ regulates the frequency decomposition at an architectural level. Changing $\alpha$ causes the network topology to change and thus can not be done during training.\\  
One of the benefits of the new feature representation is the reduction of the spatial redundancy and the  compactness compared with the original representation. Furthermore, octave convolution enable efficient communication between the high and the low frequency component of the feature representation.

\section{\uppercase{Method}}
\noindent In the following section, we describe our approach which addresses the derivation and integration of the soft octave convolution in GANs.

\subsection{Soft Octave Convolution}
\noindent First experiments with octave convolutions (see Figs.\,\ref{fig:all}, \ref{fig:evo} and \ref{fig:res}) showed that frequency factorization appears to make GAN training more efficient and stable. However, in the same  experiments we also observed that it is quite hard to pick the best value for the hyper-parameter $\alpha$: While shifting towards more lower frequent convolutions, the GAN training becomes more stable, while at the same time the lack of high frequencies makes the generated images blurry. \\
To overcome this trade off, we suggest a novel re-formulation of the octave convolutions which allows to change the ratio of the frequency factorization during training: the soft octave convolution. Instead of using a fixed factorization ration $\alpha$, we introduce two independent\footnote{In special cases $\beta_L$ and $\beta_H$ can be coupled like $\beta_L=1-\beta_H$.} ratio factors

\begin{align}
\begin{split}
	\beta_{\mathrm{L}} X^{\mathrm{L}} \quad \mathrm{and} \quad \beta_{\mathrm{H}} X^{\mathrm{H}}.
\end{split}
\end{align}
\vspace{1px}

\begin{figure}[!t]
\begin{center}
   \includegraphics[width=\linewidth]{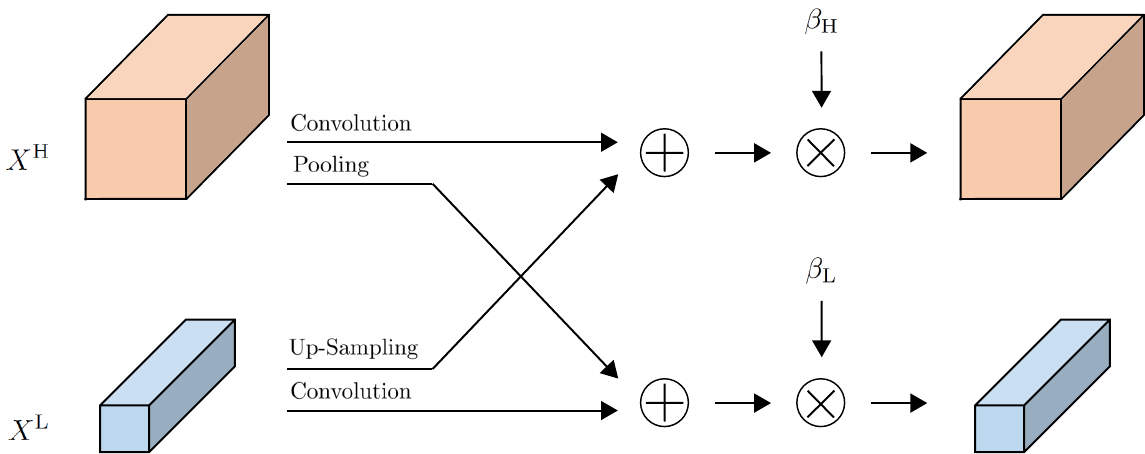}
\end{center}
   \caption{Schematic overview of the soft octave convolution. Inputs $X$ to convolutional layers are separated into  $X = \{X^{\mathrm{H}}, X^{\mathrm{L}}\}$ along the channel domain as in the original formulation. We fix $\alpha=0.5$ and introduce ratio factors $\beta_{\mathrm{L}} X^{\mathrm{L}}$ and $\beta_{\mathrm{H}} X^{\mathrm{H}}$.}
\label{fig:soft_scheme}
\end{figure}

Fig.\,\ref{fig:soft_scheme} shows a schematic overview of our soft octave convolutions.
Setting a fixed $\alpha=0.5$, we use the $\beta$ weights to shift the ration between high and low frequencies. This allows us to to apply different training schedules, e.g. forcing the GAN to learn the low frequency parts of an image before the high frequent details (see Fig.\,\ref{fig:betas} for example ratio schedules). The results in the experimental section show, that this way, GANs with soft octave convolutions are even more stable and produce high detail images.     

\vspace{1px}

\subsection{Model Architecture}

\noindent We use common GAN architectures and simply replace all standard convolutions in the generator as well as in the discriminator with the original octave \cite{chen2019drop}  and our proposed soft octave convolution. Such a change has almost no consequences on the architecture elements since it has been designed in a generic way making it a plug-and-play component. However, octave convolution has some impact on batch normalization layers. This regularization technique expects to have as input the same amount of activations from the feature maps. Because of the octave convolution nature, the size of feature maps will diverge between low and high frequency maps. To cope with this issue, two independent batch normalizations will be deployed, one for the low and one for the high frequency feature maps.

\subsection{Insights on the Frequency Domain}
\noindent So far, we motivated the use of (soft) octave convolutions by the intuition, that the training process of GANs will become more stable if we force the networks to focus on the coarse (low frequency) image structures in early stages of the training, before adding (high frequent) image details later on. While our experiments confirm this assumption (see Fig.\,\ref{fig:alls}), we follow the recent findings in \cite{durall2020watch} for the theoretical analysis of this effect. It has been shown in \cite{durall2020watch}, that most up-sampling and up-convolution units, which are commonly used in GAN generators, are violating the sampling theorem. Hence, convolutional filters in the generator are prune to produce massive high frequency artifacts in the output images which can be detected by the discriminator and thus resulting in unstable training runs. While \cite{durall2020watch} propose to use larger convolutional filters and an extra regularization in order to fix this problem, soft octave convolutions allow us to regulate the spectrum of the output images directly. Fig.\,\ref{fig:soft} shows the frequency spectrum of images generated by GANs with soft octave convolutions in comparison with the vanilla case.

\begin{figure*}[!t]
\begin{subfigure}{.33\linewidth}
\centering
\includegraphics[width=\linewidth]{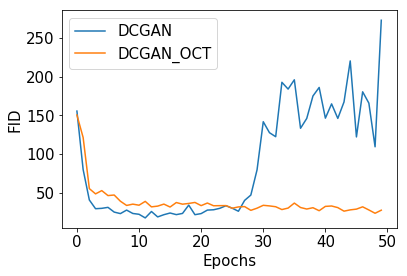}
\caption{DCGAN baseline.}
\label{fig:r}
\end{subfigure}
\begin{subfigure}{.33\linewidth}
\centering
\includegraphics[width=\linewidth]{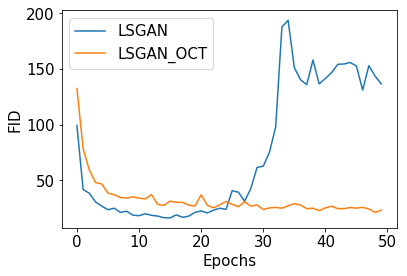}
\caption{LSGAN baseline.}
\label{fig:d}
\end{subfigure}
\begin{subfigure}{.33\linewidth}
\centering
\includegraphics[width=\linewidth]{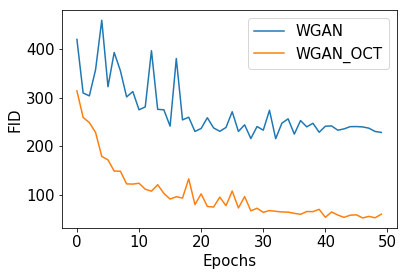}
\caption{WGAN baseline.}
\label{fig:g}
\end{subfigure}
\caption{Each figure shows the FID evolution during the training using a certain GAN implementation and its standard octave convolution variant with $\alpha=0.5$.}
\label{fig:all}
\end{figure*}

\begin{figure*}[!t]
\begin{center}
   \includegraphics[width=.9\linewidth]{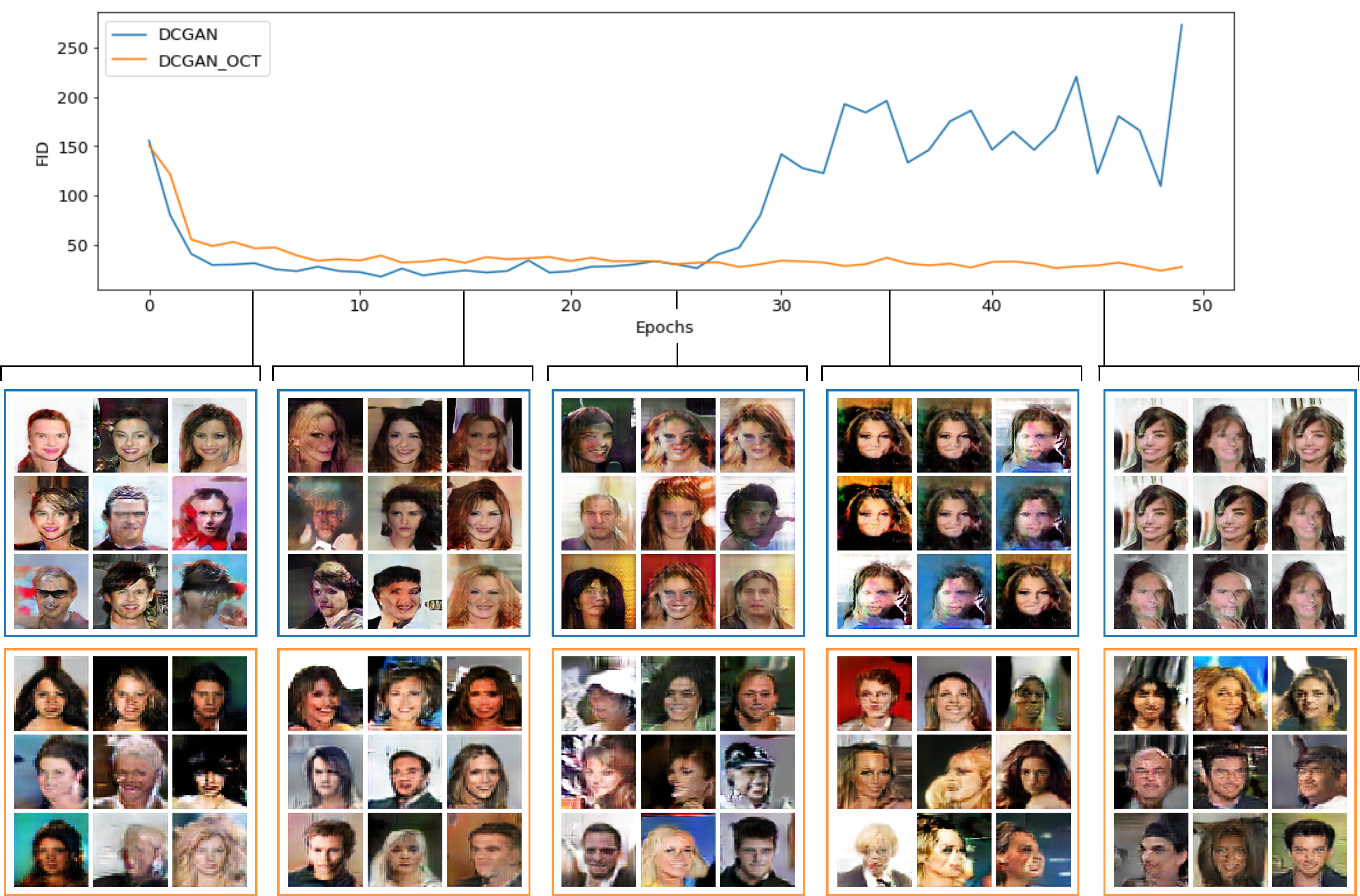}
\end{center}
   \caption{The figure shows the FID evolution together with some random generated examples using a standard DCGAN and its original octave ($\alpha=0.5$) implementation (DCGAN\_OCT).}
\label{fig:evo}
\end{figure*}

\begin{figure*}[!t]
\begin{subfigure}{.49\linewidth}
\centering
\includegraphics[width=\linewidth]{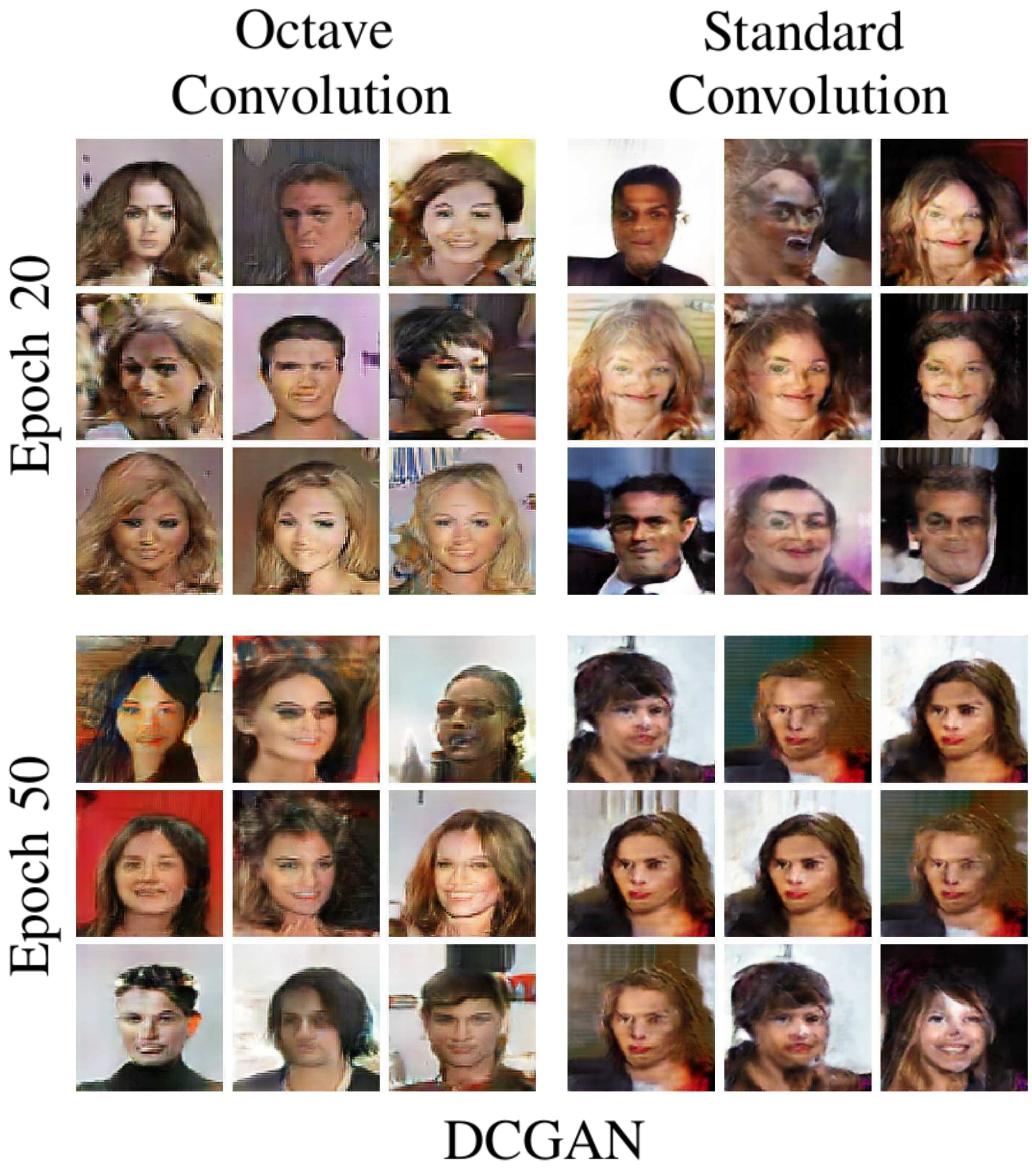}
\caption{DCGAN baseline.}
\label{fig:r1}
\end{subfigure}
\begin{subfigure}{.49\linewidth}
\centering
\includegraphics[width=\linewidth]{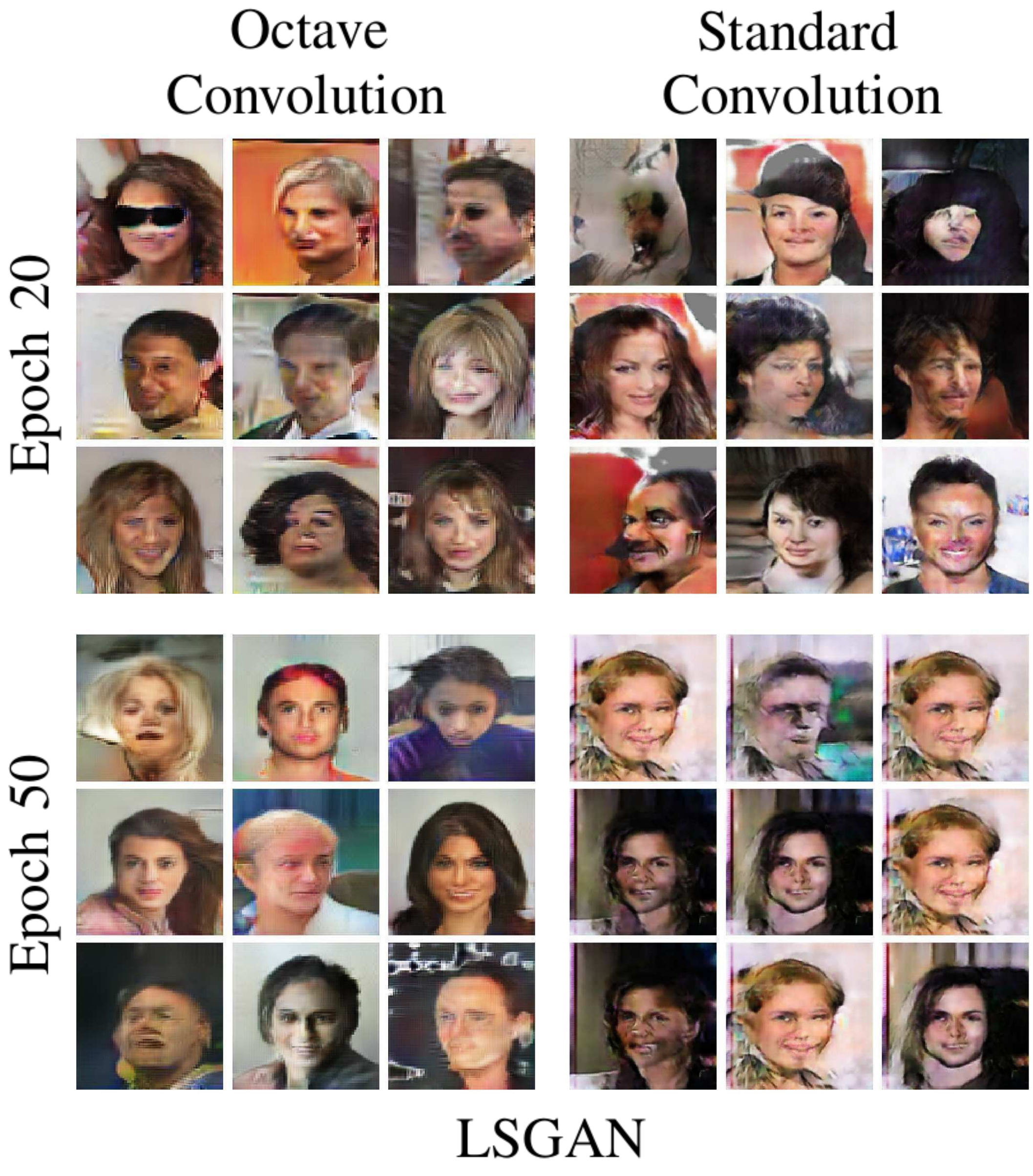}
\caption{LSGAN baseline.}
\label{fig:d2}
\end{subfigure}
\par\bigskip
\centering
\begin{subfigure}{.49\linewidth}
\centering
\includegraphics[width=\linewidth]{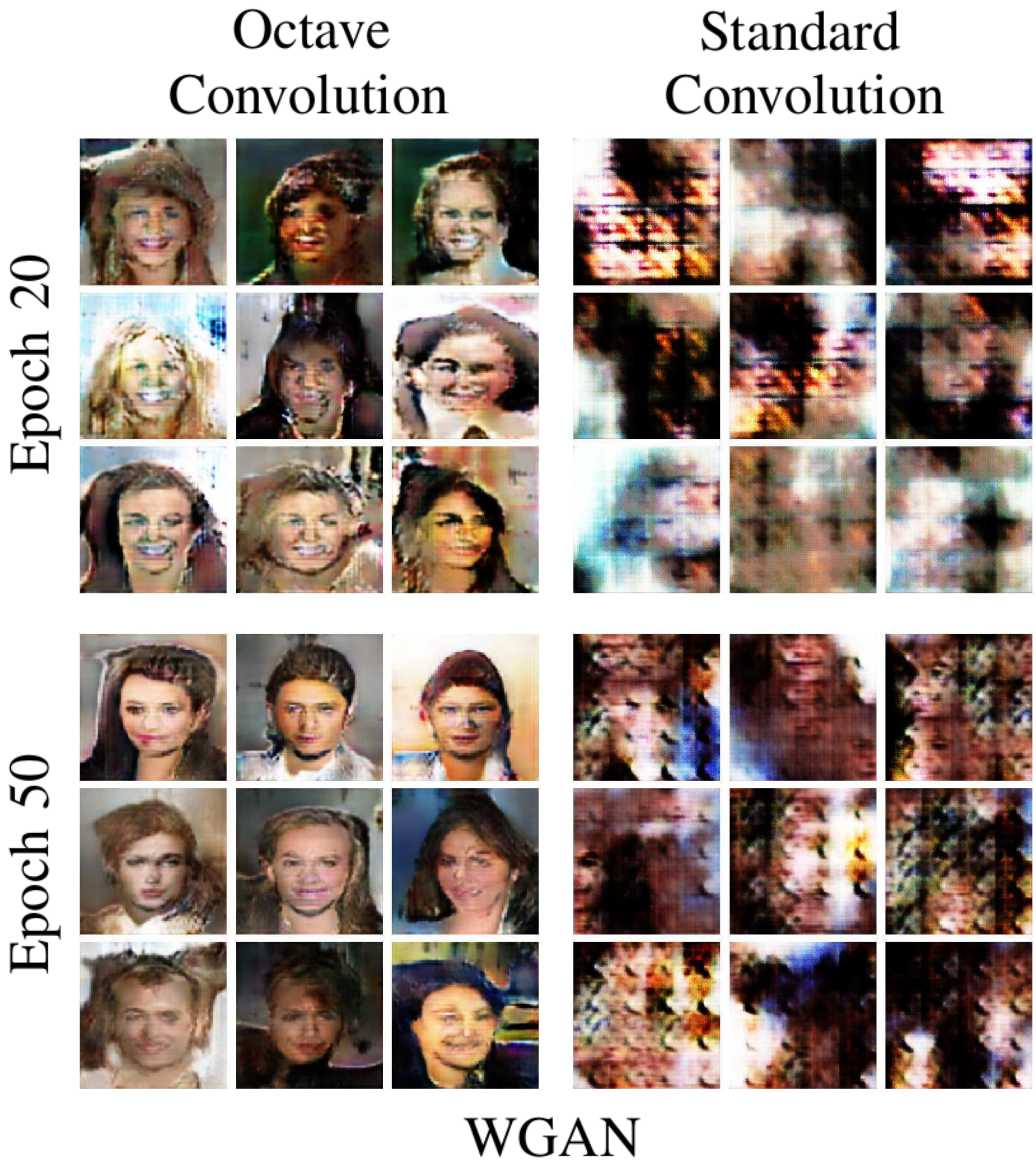}
\caption{WGAN baseline.}
\label{fig:g3}
\end{subfigure}
\caption{The figures show three independent baselines used in the experiments. Each case contains several samples across two dimensions components: the horizontal and the vertical. The first refers to the type of convolution implemented (original octave or vanilla convolution), and the second represents the stage of the training (epoch) in the particular baseline.}
\label{fig:res}
\end{figure*}

\section{\uppercase{Experiments}}
\noindent In this section, we present results for a series of experiments evaluating the effectiveness and efficiency of proposed soft octave convolutions. We first give a detailed introduction of the experimental setup. Then, we discuss the results on several different GAN architectures, and finally we explore different configurations modifying the weight of low and high frequency feature maps accordingly. Source code is available on \textit{Github}\footnote{ Source code: https://github.com/cc-hpc-itwm/Stabilizing-GANs-with-Octave-Convolutions}.

\subsection{Experimental Settings}

\noindent We train all architectures on the CelebFaces Attributes (CelebA) dataset\,\cite{liu2015deep}. It consists of 202,599 celebrity face images with variations in facial attributes. In training, we crop and resize the initially 178x218 pixel image to 128x128 pixels. All experiments presented in this paper have been conducted on a single NVIDIA GeForce GTX 1080 GPU, without applying any post-processing. Our evaluation metric is Fr\'{e}chet Inception Distance (FID)\,\cite{heusel2017gans}, which uses the Inception-v3 network pre-trained on ImageNet to extract features from an intermediate layer. Then, we model the distribution of these features using a multivariate Gaussian distribution with mean $\mu$ and covariance $\Sigma$. This procedure is conducted for both real images $x$ and generated images $z$, and it can be written as

\begin{align}
\begin{split}
	\mathrm{FID}(x,z) =\,&||\mu_{x} -\mu_{z} ||_{2}^{2}+
	 Tr(\Sigma_{x}+\Sigma_{z}-2(\Sigma_{x}\Sigma_{z})^{\frac{1}{2}}).
\end{split}
\end{align}
\vspace{1px}

Lower FID is better, corresponding to more similar real and generated samples as measured by the distance between their feature distributions.

\subsection{Training}
\noindent In this subsection, we investigate the impact of replacing the standard convolution with octave convolution. We conduct a series of studies using well-known GAN baselines which we have not optimized towards the dataset since the main objective here is to verify the impact of the new convolutional scheme and not to defeat state-of-the-art score results. In particular, we constrain our experiments to three types of GANs: DCGAN, LSGAN and WGAN. All comparisons between the baseline methods and the proposals have the same training and testing setting. We use an Adam optimizer \cite{kingma2014adam} with $\beta_{1} = 0.5$, $\beta_{2} = 0.999$ during training in all the cases. We set the batch size to 64 and run the experiments for 50 epochs. We update the generator after every discriminator update, and the learning rate used in the implementation is 0.0002.

\begin{figure*}
\begin{subfigure}{.33\linewidth}
\centering
\includegraphics[width=\linewidth]{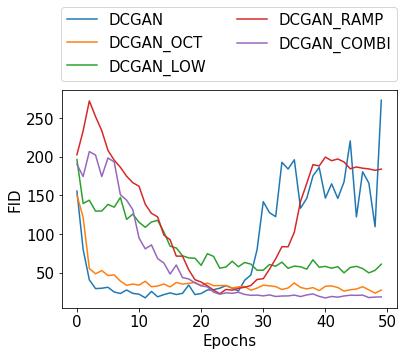}
\caption{DCGAN baseline.}
\label{fig:rs}
\end{subfigure}
\begin{subfigure}{.33\linewidth}
\centering
\includegraphics[width=\linewidth]{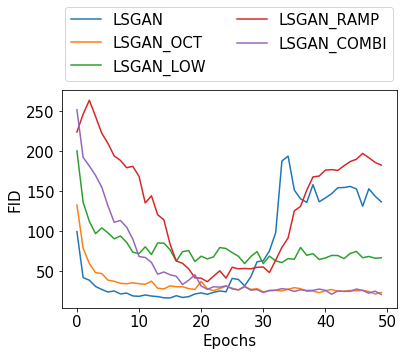}
\caption{LSGAN baseline.}
\label{fig:ds}
\end{subfigure}
\begin{subfigure}{.33\linewidth}
\centering
\includegraphics[width=\linewidth]{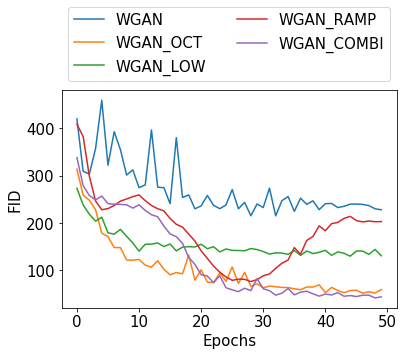}
\caption{WGAN baseline.}
\label{fig:gs}
\end{subfigure}
\caption{FID evolution during the training using different GAN implementation and their octave variants. The suffixes stand for the following: OCT octave convolution with $\alpha$ = 0.5 (vanilla configuration), LOW octave convolution with $\alpha$ = 0.99, RAMP soft octave convolution with $\alpha$ = 0.5 and $\beta$s as in\,\ref{fig:ramp}, and COMBI soft octave convolution with $\alpha$ = 0.5 and $\beta$s as in\,\ref{fig:combi}. }
\label{fig:alls}
\end{figure*}

\noindent \textbf{Standard Octave Convolution.} First, we conduct a set of experiments to validate the effect of the original octave convolution. Therefore, we set the $\alpha$ to 0.5. We begin with using the baseline models and compute the FID after each iteration. Then, we repeat the same procedure but this time we train using the octave convolution on the models. Our results in Fig.\,\ref{fig:all} show that in all three baselines, the octave model generates images of better or similar quality compared to the previous training. Moreover, we can observe the improvement of stability during training for the octave implementation.

Fig.\,\ref{fig:evo} depicts again the comparison between the vanilla DCGAN with the octave version. However, this time the plot includes an arbitrary set of samples which clearly show that these curves correlate well with the visual quality of the generated samples. Even more detailed and extended qualitative evaluations are presented in Fig.\,\ref{fig:res}, where numerous samples from all the baselines are displayed. Note that vanilla DCGAN and LSGAN start to suffer from mode collapse from epoch 25 forward. Thus, we choose epoch 20 to do a fair qualitative comparison as it seems to be the optimal training epoch. We also show the final results (epoch 50), which support the stability claim held in this work.\\

\begin{figure}[!t]
\begin{center}
   \includegraphics[width=.85\linewidth]{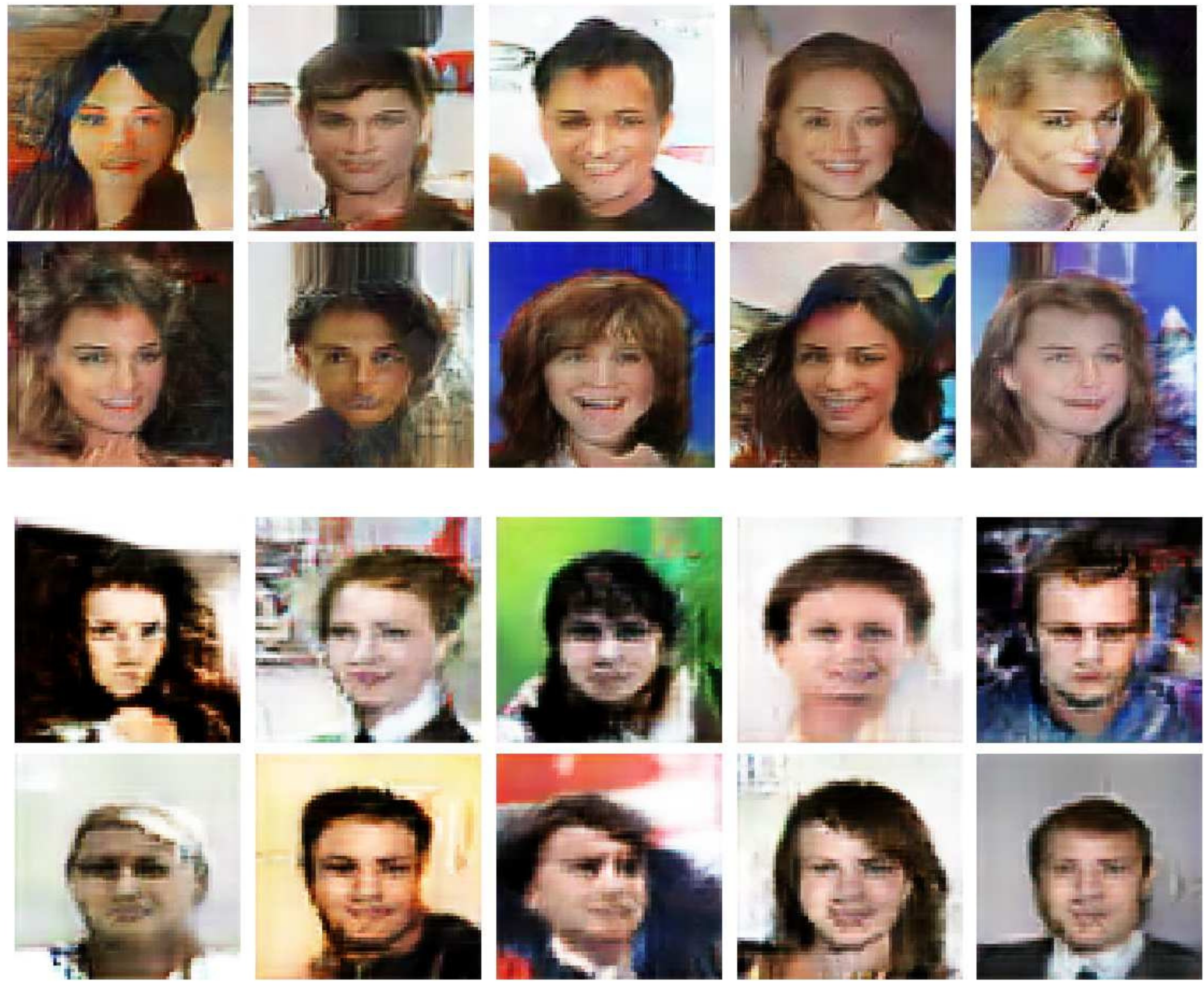}
\end{center}
   \caption{The figure compares two set of random generated images using DCGAN with octave convolution but different $\alpha$s. (Top) Implemented with $\alpha = 0.5$. (Bottom) Implemented with $\alpha = 0.99$. The bigger the $\alpha$ is, the less amount of high frequency components are present.}
\label{fig:comparation}
\end{figure}

\noindent \textbf{Soft Octave Convolutions.} In this second part of the experiments, we conduct an analysis of the impact of the low and high frequency feature maps. In order to verify how sensitive GANs are to these modifications, we start running a test for the three baselines, where we set $\alpha$ to 0.99\footnotemark
\footnotetext{We cannot set  $\alpha$ to 1 because of implementation issues. Nevertheless, the difference should be negligible.} (see Fig.\,\ref{fig:comparation}). By doing so, we get rid of all the high frequency maps, and as it is expected, the training shows constant stability since low frequencies do not contain big jumps or variations. On the other hand, surprisingly the score results are not dramatically worse than vanilla baselines (see Fig\,\ref{fig:alls}). Indeed, it is interesting to notice that both share a similar FID score evolution.

\begin{figure}[t]
\begin{subfigure}{.49\linewidth}
\centering
\includegraphics[width=1\linewidth]{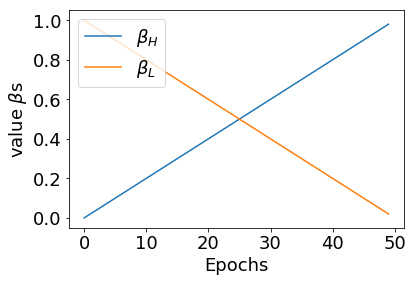}
\caption{Ramp.}
\label{fig:ramp}
\end{subfigure}
\begin{subfigure}{.49\linewidth}
\centering
\includegraphics[width=1\linewidth]{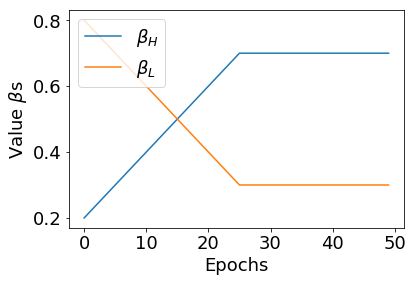}
\caption{Combination.}
\label{fig:combi}
\end{subfigure}
\caption{These plots show the weight $\beta$s curve used in ramp\,(\subref{fig:ramp}) and in combination\,(\subref{fig:combi}) scheme.}
\label{fig:betas}
\end{figure}

From the previous results, we notice the importance of  hyper-parameter $\alpha$. However, it is a well-known NP-hard problem to find the best topology in deep neural networks and in fact, it is an area of active research by itself\,\cite{elsken2018neural,liu2018darts,xie2019exploring}. As a consequence, we avoid to modify directly the topology by changing $\alpha$. Driven by these observations, finally, we conduct a new series of experiments based on two new hyper-parameters $\beta_{\mathrm{L}}$ and $\beta_{\mathrm{H}}$. Indeed, they can be seen as an extension of $\alpha$ because they will modify the feature maps too. Nonetheless, $\beta$s do not modify the amount of feature maps, but their weight. In Fig.\,\ref{fig:betas} are plotted two different strategies followed in the work. We first implement a ramp scheme (see Fig.\,\ref{fig:ramp}). The intuition behind is that low frequency signals that capture the global layout and coarse structure are learnt at the beginning, and after a certain time the high frequency parts that capture fine details, start to appear and gain more importance. Trying to capture such a behaviour, we deploy the ramp evolution. Nonetheless, this strategy might be too harsh as the role played by the low and high frequencies is too insignificant at certain training stages (see Fig.\,\ref{fig:alls}). As a result, we introduce a second weighting strategy called combination (see Fig.\,\ref{fig:combi}), which tries to be a trade-off between frequency components offering an optimized combination. In Fig.\,\ref{fig:alls} are shown the three baselines and their octave variants.

\subsection{Stability and Effects in the Frequency Domain}
\begin{figure}[t]
\begin{subfigure}{.49\linewidth}
\centering
\includegraphics[width=1\linewidth]{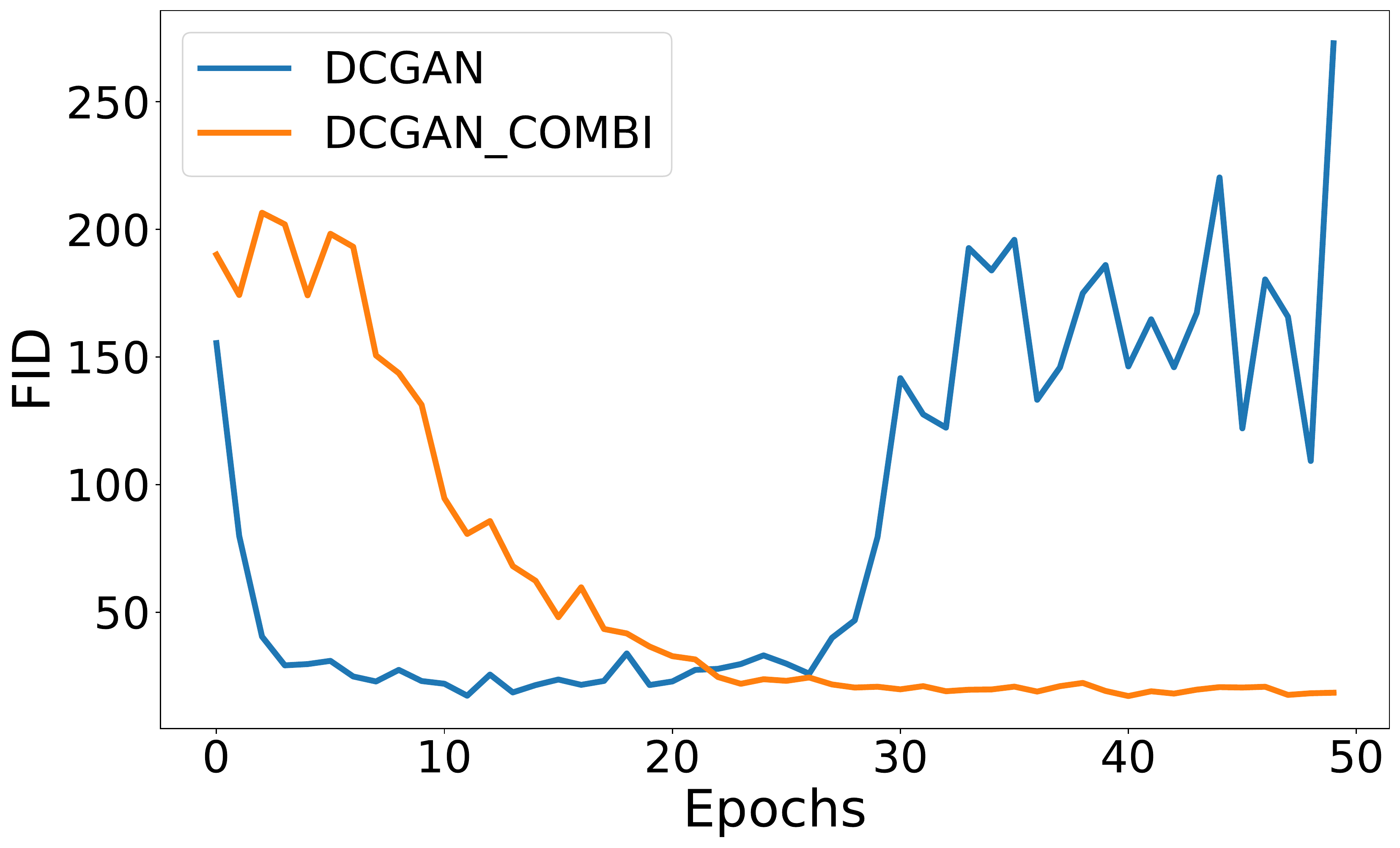}
\caption{FID evolution during the training.}
\label{fig:fid_dcgan}
\end{subfigure}
\begin{subfigure}{.49\linewidth}
\centering
\includegraphics[width=1\linewidth]{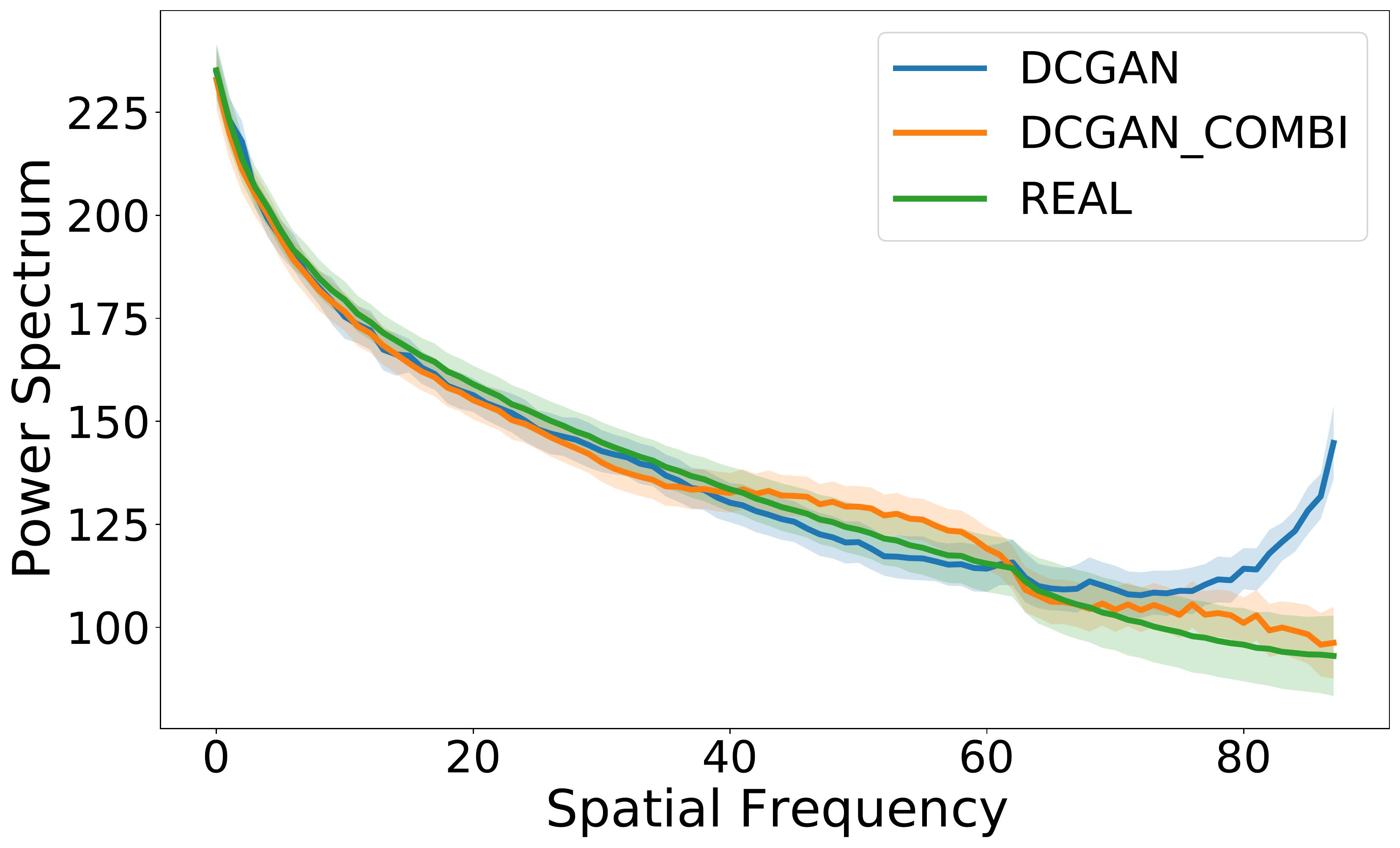}
\caption{1D power spectrum from the outputs.}
\label{fig:freq_dcgan}
\end{subfigure}
\caption{These plots depict the comparison between standard convolution and soft octave convolution in terms of stability and frequency domain. }
\label{fig:soft}
\end{figure}

\noindent Finally, in this subsection we analyse the impact of the soft octave convolution in the frequency domain. In Fig.\,\ref{fig:soft}, we  compare the standard convolution with the COMBI soft octave convolution. On the one hand, we have the FID curves that describe the stability during training. As we have seen in previous sections, our approach guarantees a much more stable behaviour without having any break. On the other hand, inspired by\,\cite{durall2019unmasking,durall2020watch}, we take the outputs from the different methods and we compute their spectral components after the training is over. This experiment allows to confirm the effect that our method has on the spectrum domain, being able to correct the artifacts that standards convolutions have in the high frequency band. In other words, soft octave convolution pushes the frequency components towards the real one.

\section{\uppercase{Conclusions}}
\noindent In this work, we tackle the common problem of GAN training stability. We propose a novel yet simple convoluitonal layer coined as soft octave convolution. Intuitively, our approach forces GANs to learn low frequency coarse image structures before descending into fine (high frequency) details. As a result, we achieve both a stable training and a reduction of common artifacts present in the high frequency domain of generated images. Furthermore, we show how this method is orthogonal and complementary to existing methods and leads to generate images of better or equal quality suppressing the mode collapse problem. We believe the line of this work opens interesting avenues for feature research, including exploring Bayesian optimizations.

\bibliographystyle{apalike}
{\small
\bibliography{example}}

\end{document}